\documentclass[runningheads]{llncs}
\usepackage[T1]{fontenc}
\usepackage{graphicx}
\usepackage{amsfonts}
\usepackage{booktabs}
\usepackage{multirow}
\usepackage{graphicx}
\usepackage{amsmath}
\usepackage{amsfonts,amssymb}
\usepackage{tikz}
\usepackage{pifont}
\usepackage{url}
\usepackage{threeparttable}
\usepackage{marvosym}
\usepackage{hyperref}
\hypersetup{
    colorlinks=true,
    linkcolor=black,
    filecolor=black,
    urlcolor=black,
    citecolor=black,
}
\begin{document}
\title{Empowering Few-Shot Relation Extraction with \\The Integration of Traditional RE Methods and\\ Large Language Models}
\titlerunning{FSRE with the Integration of Traditional RE Methods and LLMs}
%
\author{Ye Liu, Kai Zhang\textsuperscript{(\Letter)}, Aoran Gan, Linan Yue, Feng Hu,\\ Qi Liu, Enhong Chen}
\authorrunning{Y. Liu et al.}
\institute{$^1$School of Data Science, School of Computer Science and Technology, \\University of Science and Technology of China\\ $^2$State Key Laboratory of Cognitive Intelligence\\
\email{\{liuyer,gar,lnyue,fenghufh3\}@mail.ustc.edu.cn\\ 
\{kkzhang08,qiliuql,cheneh\}@ustc.edu.cn}}
%
%
\maketitle              
\pagestyle{empty}  
\thispagestyle{empty} 

\begin{abstract}
Few-Shot Relation Extraction (FSRE), a subtask of Relation Extraction (RE) that utilizes limited training instances, appeals to more researchers in Natural Language Processing (NLP) due to its capability to extract textual information in extremely low-resource scenarios. The primary methodologies employed for FSRE have been fine-tuning or prompt tuning techniques based on Pre-trained Language Models (PLMs). Recently, the emergence of Large Language Models (LLMs) has prompted numerous researchers to explore FSRE through In-Context Learning (ICL). However, there are substantial limitations associated with methods based on either traditional RE models or LLMs. Traditional RE models are hampered by a lack of necessary prior knowledge, while LLMs fall short in their task-specific capabilities for RE. To address these shortcomings, we propose a Dual-System Augmented Relation Extractor (DSARE), which synergistically combines traditional RE models with LLMs. Specifically, DSARE innovatively injects the prior knowledge of LLMs into traditional RE models, and conversely enhances LLMs' task-specific aptitude for RE through relation extraction augmentation. Moreover, an Integrated Prediction module is employed to jointly consider these two respective predictions and derive the final results. Extensive experiments demonstrate the efficacy of our proposed method.

\end{abstract}
\section{Introduction}
Relation Extraction (RE) aims to determine the relation expressed between two entities within an unstructured textual context~\cite{zhou2022improved}. Few-Shot Relation Extraction (FSRE), as a subtask of RE, seeks to solve the RE problem by utilizing only K instances per relation (K-shot) in the training and validation phases~\cite{chen2022knowprompt,xu2023unleash}. 

\begin{figure*}[t]
    \centering
    \includegraphics[width=10.5cm]{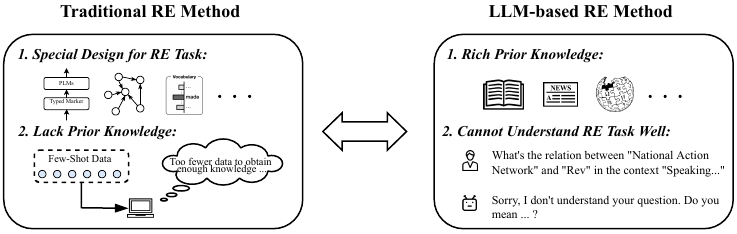}
    \caption{The comparison between traditional RE methods and LLM-based RE methods.}
    \label{fig:intro}
\end{figure*}

The primary methodologies employed to address the FSRE task have been fine-tuning or prompt tuning techniques grounded on Pre-trained Language Models (PLMs)~\cite{chen2022knowprompt,zhou2022improved}. Recently, with the emergence of Large Language Models (LLMs), numerous researchers have embarked on the exploration of FSRE through the In-Context Learning (ICL) technology~\cite{gutierrez2022thinking,wan2023gpt,xu2023unleash}. However, there are substantial limitations associated with methods based on either traditional RE models or LLMs. As depicted in Figure~\ref{fig:intro}, although most traditional RE methods are custom-built for the RE task, they still lack necessary prior knowledge that is crucial for resolving many domain-specific cases~\cite{chen2022knowprompt,han2022generative}. Acquiring such prior knowledge is particularly challenging in extremely low-resource settings, such as an 8-shot scenario. On the other hand, methods based on LLMs present a contrasting issue. With the scaling of model size and corpus size, LLMs possess an extraordinary amount of prior knowledge. Nevertheless, given that these LLMs are designed for general usage, they lack the task-specific ability for RE, which makes it difficult to fully harness their prior knowledge. This dichotomy between the strengths and weaknesses of traditional RE models and LLMs presents a novel perspective in the field of few-shot relation extraction.

To this end, this paper proposes a novel approach that amalgamates the traditional RE methods with LLMs. By doing so, we aim to address the aforementioned shortcomings by capitalizing on their respective strengths. Specifically, we develop a Dual-System Augmented Relation Extractor (DSARE) for few-shot relation extraction. DSARE consists of three key components: (a) A LLM-augmented RE module: This module designs prompts that enable LLMs to generate additional in-domain labeled data to boost the training of traditional RE models, thereby effectively injecting the prior knowledge of LLMs into the traditional RE methods. (b) A RE-augmented LLM module: This module utilizes the trained RE model to identify and retrieve the most valuable samples from the training data. These samples are subsequently employed as demonstrations for the In-Context Learning of LLMs, thereby enhancing their RE-specific aptitude. (c) An Integrated Prediction module: It takes into account the predictions of both the LLM-augmented RE and RE-augmented LLM modules. When the two predictions differ, a specially designed selector is activated to make a final decision. Finally, extensive experiments on three publicly available datasets demonstrate the effectiveness of our proposed method, and further indicate the necessity to integrate traditional RE models and LLMs.

Our code is available via \url{https://github.com/liuyeah/DSARE}.

\section{Related Work}
\subsubsection{Few-shot Relation Extraction.}
Due to the large computation ability of pre-trained language models, existing few-shot relation extraction methods mainly adopt the fine-tuning method to solve the few-shot relation extraction problem~\cite{lyu2021relation,zhou2022improved}. In recent years, in order to bridge the gap between pre-training objectives and RE task, prompt tuning has been proposed and demonstrated remarkable capability in low-resource scenarios~\cite{chen2022knowprompt,han2022generative,han2022ptr}. 

Currently, with the arise of large language models, many researchers attempt to tackle few-shot relation extraction via In-Context Learning technology~\cite{gutierrez2022thinking,wan2023gpt,xu2023unleash}. However, these approaches simply apply LLMs to few-shot relation extraction tasks through straightforward queries, which fails to fully harness the potential of LLMs. More importantly, they overlook the possibility that LLMs and traditional RE models could mutually enhance each other's performance.

\subsubsection{Large Language Models.}
The emergence of Large Language Models (LLMs) such as GPT-4, LLama-2 and others~\cite{openai2023gpt4,ouyang2022training,touvron2023llama,tunstall2023zephyr}, represents a signiﬁcant advancement in the field of natural language processing. By leveraging In-Context Learning, a novel few-shot learning paradigm was ﬁrst introduced by~\cite{brown2020language}. Up to now, LLMs have demonstrated remarkable performance across a range of NLP tasks, such as text classiﬁcation, named entity recognition, question answering and relation extraction~\cite{gutierrez2022thinking,liu2022makes,wan2023gpt,xu2023unleash}.

Previous research efforts~\cite{gutierrez2022thinking,wan2023gpt,xu2023unleash} have sought to solve few-shot relation extraction by directly asking LLMs or retrieving more suitable demonstrations. For instance, Wan et al.~\cite{wan2023gpt} attempted to introduce the label-induced reasoning logic to enrich the demonstrations. Meanwhile, Xu et al.~\cite{xu2023unleash} designed task-related instructions and a schema-constrained data generation strategy, which could boost previous RE methods to obtain state-of-the-art few-shot results.

\section{Problem Statement}
Let $C$ denote the input text and $e_{sub} \in C$, $e_{obj} \in C$ denote the pair of subject and object entities. Given the entity type of $e_{sub}$, $e_{obj}$, and a set of pre-defined relation classes $\mathbb{R}$, relation extraction aims to predict the relation $y \in \mathbb{R}$ between the pair of entities $(e_{sub}, e_{obj})$ within the context $C$~\cite{wan2023gpt,zhou2022improved}. 

As for the few-shot settings, following the strategy adopted by~\cite{gao2021making,xu2023unleash}, we randomly sample K instances per relation (K-shot) for the training and validation phases. The whole test set is preserved to ensure the effectiveness of evaluation.

\section{DSARE Model}
\subsection{LLM-augmented RE}
\label{model:RE}
\subsubsection{LLM Data Augmentation.}
In this part, we aim to implement the data augmentation via LLMs, anticipated to enrich the training data for relation extraction. Specifically, drawing inspiration from~\cite{xu2023unleash}, we construct prompts to tell the LLM the essential components of one RE training sample, i.e., context text, subject entity, object entity, subject entity type, object entity type and the relation. Then the LLM is guided to create more pseudo RE samples. Upon receiving the outputs from the LLM, we establish rules, such as regular expressions, to transform the augmented RE data into the desired format. 

\begin{figure*}[t]
    \centering
    \includegraphics[width=12cm]{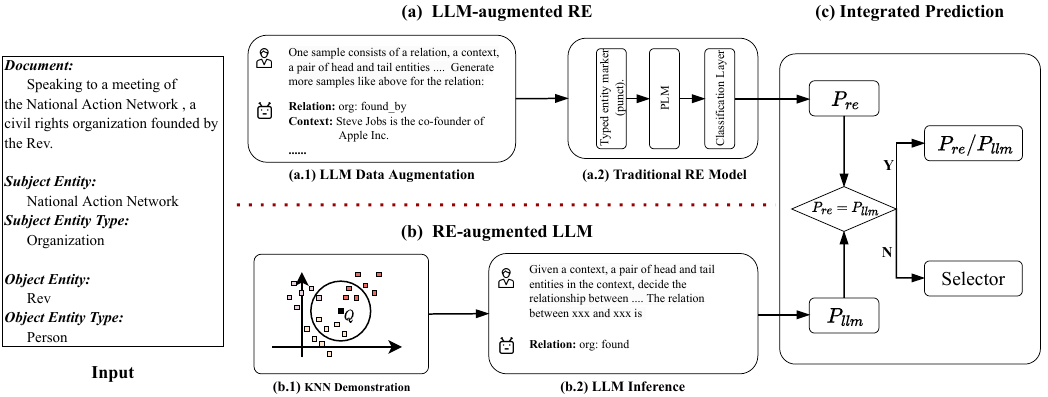}
    \caption{The architecture of DSARE. It includes three parts: (a) LLM-augmented RE module; (b) RE-augmented LLM module; (c) Integrated Prediction module.}
    \label{fig:archirecture}
\end{figure*}

\subsubsection{Traditional RE Model.}
With the augmented datasets, we obtain more diverse data to train a traditional RE model. Here we adopt the Typed Entity Marker (punct) method proposed by~\cite{zhou2022improved} to denote the entity and context text, and further train a relation extraction model. Specifically, we utilize the symbols “@” and “\#” to denote the start/end of the subject and object entities, and further adopt the symbols “$\ast$” and “$\wedge$” to indicate the subject and object entity types, respectively. The processed text is then fed into the pre-trained language model to obtain the representations of the subject and object entities ($h_{sub}$, $h_{obj}$) via the special token “@” and “\#”. Finally, we pass ($h_{sub}$, $h_{obj}$) into a classification layer to derive the results.

\subsection{RE-augmented LLM}
\label{model:LLM}
\subsubsection{KNN Demonstration.}
In Section~\ref{model:RE}, we train a traditional relation extraction model, which allows us to implement a k-nearest neighbors (KNN) search method to retrieve more valuable samples from the training set. Specifically, we utilize the obtained entity representation $H = [h_{sub}, h_{obj}]$ to represent each sample, and further obtain the representation and label pair $(H_i, r_i)$ on the training set, which we denote as a datastore $D$.

When inferring a new sample $j$, we utilize its entity representation $H_j$ to query $D$ according to the euclidean distance to obtain the $k$ nearest neighbors: $\mathcal{N} = \{ (H_i, r_i) \}_{i=1}^k$, which we adopt as demonstrations for LLM inference.

\subsubsection{LLM Inference.}
After obtaining the effective demonstrations, we design prompts to provide the essential information to the LLM, thus generating the LLM results. Specifically, inspired by the various attempts about ICL~\cite{xu2023unleash}, we first describe the target of the relation extraction task through a instruction. Then, the retrieved $k$ nearest neighbors $\mathcal{N} = \{ (H_i, r_i) \}_{i=1}^k$ of current sample are followed, which provide the most relevant information to the LLM. Finally, we ask the LLM to predict the relation of current sample.

\subsection{Integrated Prediction}
\label{sec:fusion}
In Section~\ref{model:RE} and \ref{model:LLM}, we apply traditional RE models and LLMs to conduct few-shot relation extraction from dual perspectives. In this part, we aim to obtain the final outputs by considering both the LLM-augmented RE inference result $P_{re}$ and the RE-augmented LLM inference result $P_{llm}$.

More specifically, as illustrated in Figure~\ref{fig:archirecture}, if the two results are equal (i.e., $P_{re} = P_{llm}$), our model directly yields the predicted relation. In circumstances where the two results diverge, we design a selector to further ask the LLM to make a choice between these two relations. In order to improve the effectiveness of the selector, we directly retrieve $m$ samples labeled with these two relations from the training dataset, respectively. Subsequently, we ask the LLM via a similar way we introduced in \textbf{LLM Inference} to obtain the final results\footnote{If the LLM does not make an inference or we are unable to convert the output into the pre-defined relations, we will conclude there is no relation between subject and object entities. Note that $no\_relation$ is also a relation category in these datasets.}. 

\section{Experiments}
\subsection{Experimental Setup}
\label{sec:exp_setup}

\subsubsection{Datasets and Evaluation Metrics.} 
For extensive experiments, we conduct our experiments on three widely-used relation extraction datasets: TACRED~\cite{zhang2017position}, TACREV~\cite{alt2020tacred} and Re-TACRED~\cite{stoica2021re}. More statistics about the datasets can be found in Table~\ref{tab:data}. Regarding the evaluation metrics, we adopt the micro-F1 scores of RE as the primary metric to evaluate models, considering that F1 scores can assess the overall performance of precision and recall~\cite{chen2022knowprompt,liu2023techpat,liu2023enhancing,zhang2022incorporating}.

\begin{table}[t]
    \centering
    \caption{Data Statistics}
    \renewcommand\arraystretch{0.9}
    \setlength{\tabcolsep}{4mm}{
    \begin{tabular}{c|c|c|c|c}
    \toprule
        Dataset & \textbf{\#Train} & \textbf{\#Dev} & \textbf{\#Test} & \textbf{\#Rel} \\
        \midrule
        TACRED & 8/16/32 & 8/16/32 & 15,509 & 42\\
        TACREV & 8/16/32 & 8/16/32 & 15,509 & 42\\
        Re-TACRED & 8/16/32 & 8/16/32 & 13,418 & 40\\
    \bottomrule
    \end{tabular}}
    \label{tab:data}
\end{table}

\subsubsection{Implementation Details.}
In this paper, we utilize the \textit{zephyr-7b-alpha}~\cite{tunstall2023zephyr} model on Huggingface as the LLM to conduct experiments. 

In the Traditional RE Model part (Section~\ref{model:RE}), we adopt \textit{roberta-large}~\cite{liu2019roberta} as the base architecture. The batch size is set to 4, and the model is optimized by AdamW~\cite{loshchilov2017decoupled} with a learning rate of $3e-5$. We train the model on the training set for 50 epochs and choose the best epoch based on the micro-F1 performance on the development set.

In the LLM Data Augmentation part (Section~\ref{model:RE}), we double the K-shot training set through LLMs. That is to say, for an 8-shot training set, we construct 8 pieces of pseudo data per relation, thereby creating the final augmented training set. In the KNN Demonstration part (Section~\ref{model:LLM}), we set the number of retrieved nearest neighbors as $k=8$. In the Integrated Prediction module (Section~\ref{sec:fusion}), we set the number of retrieved samples for each relation as $m=4$.

\subsubsection{Benchmark Methods.}
We compare our methods with the state-of-the-art few-shot relation extraction methods. According to the modeling architecture, they can be grouped into three categories, including Traditional RE Methods (\ding{172} $\sim$ \ding{175}), LLM-based Methods (\ding{176} $\sim$ \ding{178}) and Hybrid Methods (\ding{179}).

\begin{itemize}
    \item \textbf{\ding{172} TYP Marker}~\cite{zhou2022improved} proposes to incorporate entity representations with typed markers, which presents remarkable performance on the RE task.
    \item \textbf{\ding{173} PTR}~\cite{han2022ptr} designs prompt tuning with rules for relation extraction tasks and applies logic rules to construct prompts with several sub-prompts.
    \item \textbf{\ding{174} KnowPrompt}~\cite{chen2022knowprompt} innovatively injects the latent knowledge contained in relation labels into prompt construction with the learnable virtual type words and answer words.
    \item \textbf{\ding{175} GenPT}~\cite{han2022generative} proposes a novel generative prompt tuning method to reformulate relation classiﬁcation as an inﬁlling problem, which exploits rich semantics of entity and relation types.
    \item \textbf{\ding{176} GPT-3.5}~\cite{ouyang2022training}, \textbf{\ding{177} LLama-2}~\cite{touvron2023llama}, \textbf{\ding{178} Zephyr}~\cite{tunstall2023zephyr} are the advanced LLMs. We leverage the API for GPT-3.5, while adopt the 7B version for LLama-2 (llama-2-7b-chat-hf) and Zephyr (zephyr-7b-alpha). We utlize the prompt from~\cite{xu2023unleash} to conduct In-Context Learning.
    \item \textbf{\ding{179} Unleash}~\cite{xu2023unleash} proposes schema-constrained data generation methods\footnote{For fair comparison, we apply this data generation method to double the training set, which is the same as our settings introduced in the Implementation Details part.} through LLMs, which boost previous RE methods (i.e., KnowPrompt) to obtain more competitive results.
\end{itemize}

\begin{table*}[t]
    \centering
    \caption{Experimental Results (\%)}
    \begin{threeparttable}
    \renewcommand\arraystretch{0.9}
    \setlength{\tabcolsep}{1.1mm}{
    \begin{tabular}{ l|c|c|c|c|c|c|c|c|c }
    \toprule
    \multirow{2}{*}{Methods}&
    \multicolumn{3}{c}{TACRED}&
    \multicolumn{3}{|c}{TACREV} & \multicolumn{3}{|c}{Re-TACRED} \\
      & K=8 & K=16 & K=32 & K=8 & K=16 & K=32 & K=8 & K=16 & K=32 \\
    \midrule
    \ding{172} TYP Marker  & 29.02 & 31.35 & 31.86 & 26.28 & 29.24 & 31.55 & 51.32 & 55.60 & 57.82\\
    \ding{173} PTR  & 28.34 & 29.39 & 30.45 & 28.63 & 29.75 & 30.79 & 47.80 & 53.83 & 60.99\\
    \ding{174} KnowPrompt  & 30.30 & 33.53 & 34.42 & 30.47 & 33.54 & 33.86 & 56.74 & 61.90 & 65.92\\
    \ding{175} GenPT  & 35.45 & 35.58 & 35.61 & 33.81 & 33.93 & 36.72 & 57.03 & 57.66 & 65.25\\
    \midrule
    \ding{176} GPT-3.5  & \multicolumn{3}{c|}{29.72} & \multicolumn{3}{c|}{29.98} & \multicolumn{3}{c}{39.06} \\
    \ding{177} LLama-2  & \multicolumn{3}{c|}{22.68} & \multicolumn{3}{c|}{21.96} & \multicolumn{3}{c}{34.31} \\
    \ding{178} Zephyr  & \multicolumn{3}{c|}{37.10} & \multicolumn{3}{c|}{38.83} & \multicolumn{3}{c}{35.81} \\
    \midrule
    \ding{179} Unleash  & 32.24 & 33.81 & 34.76 & 32.70 & 34.53 & 35.28 & 58.29 & 64.37 & 66.03\\
    \midrule
    \textbf{DSARE (ours)}  & \textbf{43.84} & \textbf{45.40} & \textbf{45.94} & \textbf{44.69} & \textbf{46.61} & \textbf{46.94} & \textbf{60.04} & \textbf{66.83} & \textbf{67.13}\\
    
    \bottomrule
    \end{tabular}
    }
    \label{tab:main_result}
    \end{threeparttable}
\end{table*}

It is worth noting that, for these LLM-based baselines (\ding{176} $\sim$ \ding{178}), due to the limitations of maximum tokens and the fact that these datasets have at least 40 relations, we utilize the one-shot demonstration per relation following the strategy proposed by~\cite{xu2023unleash}. In contrast, our DSARE method, as introduced in the Implementation Details part, requires a maximum of 16 demonstrations\footnote{In the KNN Demonstration part (Section~\ref{model:LLM}), the number of retrieved nearest neighbors is $k=8$. And in the Integrated Prediction module (Section~\ref{sec:fusion}), we need a maximum of $2m = 8$ additional demonstrations.}, which is much fewer than the number of the one-shot demonstration per relation setting ($>=40$), thus avoiding unfair comparison.

\subsection{Experimental Result}
The main results are illustrated in Table~\ref{tab:main_result}. Our proposed DSARE model outperforms all baselines across all metrics. Particularly on the TACRED and TACREV datasets, our method manifests a significant advantage. This demonstrates the effectiveness of our designs and the benefits of integrating traditional RE models and LLMs. Furthermore, there are also some interesting phenomena:

First, the vast majority of methods exhibit superior performance on the Re-TACRED dataset compared to the TACRED and TACREV datasets. This is reasonable as Re-TACRED is an improved version among these three datasets, which addresses some shortcomings of the original TACRED dataset, refactors its training set, development set and test set. The more precise labels contribute to the learning process of these models, thereby yielding superior performance. Second, among these LLM-based methods, Zephyr (7B) demonstrates competitive performance and significantly outperforms GPT-3.5 and LLama-2 on the TACRED and TACREV datasets. This proves its strong information extraction ability, as claimed in~\cite{tunstall2023zephyr}. Third, Unleash introduces a schema-constrained data augmentation method through LLMs to enhance the Knowprompt baselines. It achieves a certain degree of improvement compared to Knowprompt, verifying the the feasibility of this line of thinking. And our DSARE model significantly surpasses Unleash, which further demonstrates the effectiveness of our designs from another perspective.

\subsection{Ablation Study}

\begin{table*}[t]
    \centering
    \caption{Ablation Experiments (\%)}
    \renewcommand\arraystretch{0.9}
    \setlength{\tabcolsep}{3mm}{
    \begin{tabular}{ l|c|c|c}
    \toprule
    \multirow{2}{*}{Ablation Models}&
    \multicolumn{3}{c}{Re-TACRED} \\
      & K=8 & K=16 & K=32 \\
    \midrule
    DSARE  & \textbf{60.04} & \textbf{66.83} & \textbf{67.13} \\
    \midrule
    LLM-augmented RE  & 52.53 & 58.01 & 58.56 \\
    RE-augmented LLM  & 56.38 & 64.85 & 66.03 \\
    \midrule
    Pure RE   & 51.32 & 55.60 & 57.82 \\
    \cline{2-4}
    Pure LLM   & \multicolumn{3}{c}{35.81}  \\
    \bottomrule
    \end{tabular}
    }
    \label{tab:ablation}
\end{table*}

In this subsection, we carry out ablation experiments to validate the effectiveness of various components of DSARE model. Specifically, we first remove the Integrated Prediction module, consequently leading to two ablated variants: LLM-augmented RE and RE-augmented LLM. As shown in Table~\ref{tab:ablation}, there are obvious decreases between DSARE and its two variants, demonstrating the efficacy of the Integrated Prediction module. 

Subsequently, we further remove the LLM Data Augmentation part from LLM-augmented RE and remove the KNN Demonstration part from RE-augmented LLM. This yields two other variants, i.e., Pure RE and Pure LLM\footnote{Note that here Pure LLM is equivalent to the baseline \ding{178} Zephyr.}. From the results, both these variants perform inferiorly, especially the Pure LLM. These findings futher demonstrate the validity and non-redundancy of our designs.

\subsection{Case Study}

\begin{figure}[t]
    \centering
    \includegraphics[width=12cm]{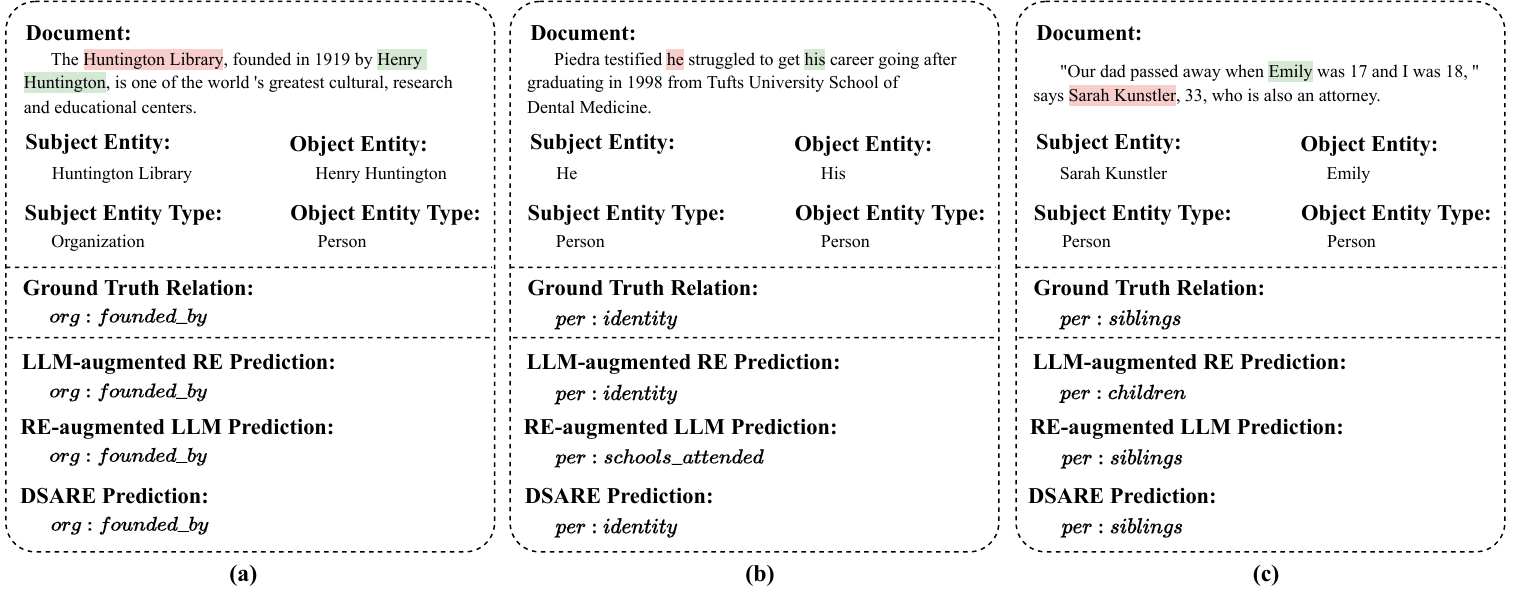}
    \caption{The case study of the DSARE model. (a) is from the TACRED dataset (K=8), while (b) and (c) are from the Re-TACRED dataset (K=8).}
    \label{fig:case}
\end{figure}

In this section, we conduct case study to more intuitively illustrate the effectiveness of integrating traditional RE models and LLMs. Specifically, as illustrated in Figure~\ref{fig:case}, we present the input information (i.e., document, subject/object entity, subject/object entity type), ground truth relation and the prediction of DSARE and its ablated variants, respectively. 

In Figure~\ref{fig:case} (a), both the LLM-augmented RE and the RE-augmented LLM make the correct prediction. In Figure~\ref{fig:case} (b) and (c), the LLM-augmented RE and RE-augmented LLM correctly infer the relations ($per: identity$ and $per: siblings$), respectively. And with the aid of the Integrated Prediction module, DSARE finally derives the correct predictions. These cases intuitively demonstrate the significant role of integrating traditional RE methods and LLMs, and further verify the validity of our DSARE model.

\section{Conclusions}
In this paper, we explored a motivated direction for empowering few-shot relation extraction with the integration of traditional RE models and LLMs. We first analyzed the necessity to joint utilize traditional RE models and LLMs, and further proposed a Dual-System Augmented Relation Extractor (DSARE). Specifically, we designed a LLM-augmented RE module, which could inject the prior knowledge of LLMs into the traditional RE models. Subsequently, a RE-augmented LLM module was proposed to identify and retrieve the most valuable samples from the training data, which provided more useful demonstrations for the In-Context Learning of LLMs. More importantly, we designed an Integrated Prediction module to joint consider the predictions of both LLM-augmented RE and RE-augmented LLM modules, thus taking advantages of each other's strengths and deriving the final results. Finally, extensive experiments on three publicly available datasets demonstrated the effectiveness of our proposed method. We hope our work could lead to more future studies.

\subsubsection{Acknowledgement.} This research was partially supported by grants from the National Natural Science Foundation of China (No. U20A20229), the Anhui Provincial Natural Science Foundation of China (No. 2308085QF229 and 2308085MG226) and the Fundamental Research Funds for the Central Universities.

%
%
%
%
\bibliographystyle{splncs04}
\bibliography{custom}

\end{document}